\documentclass[a4paper]{article}
\usepackage{INTERSPEECH2016}
\ninept

\usepackage{graphicx}
\usepackage{amssymb,amsmath,bm}
\usepackage{textcomp}
\usepackage{xcolor}
\usepackage{subfig}
\usepackage{algorithm}
\usepackage{algpseudocode}
\usepackage{multirow}
\usepackage{pifont}
\usepackage{placeins}

\ninept
\title{On the efficient representation and execution of deep acoustic models}

\makeatletter
\def\name#1{\gdef\@name{#1\\}}
\makeatother \name{{\em Raziel Alvarez, Rohit Prabhavalkar, Anton Bakhtin}\thanks{The authors would like to
    thank Peter Warden for helpful discussions regarding quantization, and Ha{\c{s}}im Sak and Kanishka Rao for helpful comments on quantization aware training in acoustic models.}}

\address{Speech Group, Google Inc. \\
  {\small \tt \{raziel,prabhavalkar,bakhtin\}@google.com}
}

\begin{document}

  \maketitle
  \begin{abstract}
In this paper we present a simple and computationally efficient quantization scheme that enables us to reduce the resolution of the parameters of a neural network from 32-bit floating point values to 8-bit integer values. The proposed quantization scheme leads to significant memory savings and enables the use of optimized hardware instructions for integer arithmetic, thus significantly reducing the cost of inference. Finally, we propose a `quantization aware' training process that applies the proposed scheme during network training and find that it allows us to recover most of the loss in accuracy introduced by quantization. We validate the proposed techniques by applying them to a long short-term memory-based acoustic model on an open-ended large vocabulary speech recognition task.

  \end{abstract}

  \section{Introduction}
  \label{sec:intro}
The use of deep learning models in software applications
and systems has experienced significant growth in the last few years.
For mobile applications, it is commonplace for these systems to make use of
powerful servers that host and execute such models.
However, there have been significant efforts~\cite{lei2013, mcgraw16, szegedy15, alsharif15}
in creating systems that can run entirely on a mobile device, which are more reliable and
have lower latency than their server-based counterparts. Such systems must be
equally accurate, while also being extremely efficient in order to avoid
consuming the limited memory and computational resources available.

In this paper, we approach creating a compact representation of neural network (NN)
models by means of a simple quantization scheme that transforms the parameters
in the model from their 32-bit floating point representation into a
lower-precision as 8-bit integers. Aside from the memory reduction resulting in
more efficient access and caching of values, the 8-bit representation allows
us to take advantage of single instruction, multiple data (SIMD) optimized
hardware instructions for integer arithmetic~\cite{Ozturk12, Cortex2012},
which are now ubiquitous in mobile devices and graphical processing units.
Thus, by performing the bulk of the operations in integer
form we significantly speed up neural network inference relative to a pure
floating point implementation, reducing latency, and power consumption.

We focus on improving embedded automatic speech recognition
system (ASR) performance following our previous work~\cite{mcgraw16},
which uses a long short term memory (LSTM) based acoustic model.
The acoustic model represents a core component that significantly
impacts final recognition accuracy, and consumes most of the
computational resources available to the system. Thus, there is a significant reward
in using more compact representations, that are also fast to execute, as long as
they do not introduce significant loss in accuracy. While we focus on LSTM
based acoustic modelling, the techniques presented can be
applied to other deep learning models and to other domains, e.g., the
text-to-speech system described in~\cite{Heiga16} uses the quantization scheme
proposed in this paper.

In Section~\ref{sec:related-work}, we
review previous work on techniques for compressing neural networks via
parameter quantization. In section~\ref{sec:deepq} we describe our proposed
quantization scheme, applied during inference and training. We describe our
experimental setup in Section~\ref{sec:experiments-setup}, and examine the
effectiveness of proposed techniques in Sections~\ref{sec:experiments}
and~\ref{sec:results}. Finally, we conclude with a discussion of our findings
in Section~\ref{sec:conclusions}.

  \section{Related work}
  \label{sec:related-work}
Neural network quantization, as well as its application during training, have been
explored as a way to represent parameters and execute inference more
efficiently and with minimal accuracy loss. The analysis of the effects of
quantization on feed-forward deep neural networks has been a field of
research for many years~\cite{Xie92, Dundar95, vanhoucke2011}. It has
established that it is indeed possible to reduce the resolution of the trained
weights from their original 32-bits without significantly affecting the network's
inference capabilities. For example, D{\"u}ndar et al.~\cite{Dundar95}, found
that a minimal resolution of 10 bits was necessary. In this paper we propose the
use of a simple uniform linear quantizer to an 8-bit resolution as
in~\cite{vanhoucke2011}.

More recently, however, there has been growing interest in incorporating 
quantization into the training procedure to account for the
noise caused by the lower resolution and precision in inference computations.
Previous work has explored reducing the resolution even further~\cite{HanMD15},
up to single bit representations
~\cite{Hwang14, Seide14, KimS16, GuptaAGN15, CourbariauxBD15, WuLWHC15}, and
extending the types of topologies from fully-connected to convolutional neural
networks (CNNs) ~\cite{HanMD15, GuptaAGN15, CourbariauxBD15, SungSH15}.

In this work, we incorporate quantization at training time to eliminate the loss
experienced when quantization is applied only \emph{after} training. Our approach is
inline with previous work~\cite{Hwang14, KimS16, CourbariauxBD15} in that
we use backpropagation with stochastic gradient descent (SGD) that performs
forward passes using quantized parameters, but utilize high precision values for
adjusting them. Unlike the previously mentioned work that focuses on reducing
resolution, we target a higher 8-bit resolution since we seek to eliminate loss
in system accuracy. Also, unlike~\cite{Hwang14, KimS16} we do
not require additional pre-training in order to initialize quantized training. Our
training scheme is applied to LSTM layers, and has also been successfully used
with CNN layers (though we do not report results in this paper). Finally,
we investigate the effects of quantization in relation to parameter
reduction, including the linear recurrent projection layer introduced into the
LSTM architecture by Sak et al.~\cite{SakSeniorBeaufays14}.



  \section{Quantization Scheme}
  \label{sec:deepq}
Given floating point values, our goal is to represent them as 8-bit integers.
Generally speaking we could use either a uniform or a non-uniform quantizer, or
even an optimal quantizer for a given value distribution~\cite{bovik2005}.
However, with simplicity and performance in mind, and validated by previous work
in the area~\cite{vanhoucke2011}, we settled on a uniform linear quantizer that
assumes a uniform distribution of the values within a given range. As a result
we do not need a decompression step at inference time.


\textbf{Quantizing.}
Given a set of float values $\mathbf{V}=\{V_x\}$, and a desired scale $S$
(e.g. we use 255 for 8-bits), we compute a factor $Q$ that produces a
scaled version of the original: $\mathbf{V}'=\{0 \le V_{x}' \le S\}$.
To maximize the use of $S$ we determine the range of values
$R = (V_\text{max} - V_\text{min})$ we want to quantize, i.e., squeeze into the
new scale $S$. Thus, the quantization factor can be expressed as $Q = \frac{S}{R}$,
and the quantized values as $V_{x}' = Q * (V_{x} - V_\text{min})$.

\textbf{Recovery.}
A quantized value $V_{x}'$ can be recovered (i.e. transformed back into its
approximate high-precision value) by performing the inverse of the quantization
operation, which implies computing a recovery factor $Q^{-1} = \frac{R}{S}$.
The recovered value is then expressed as $V_{x} = V_{x}' * Q^{-1} + V_\text{min}$.

\textbf{Quantization error and bias.}
Quantization is a lossy process, with two sources of error. The first is the difference
between the input value and its quantized-then-recovered value (precision loss).
The second is the result of discrepancies in quantization-recovery operations
that introduce a bias in the computed value (bias error)~\cite{Tan08}. Of the two sources
of error, the precision loss is theoretically and practically unavoidable but, on average,
has a smaller impact on what the original data represents (e.g. the difference in
the variances of $\mathbf{V}$ and $\mathbf{V}'$ is very small~\cite{gersho1992}). The bias, however,
is theoretically avoidable by paying close attention to how the quantized values are
manipulated, thus avoiding the introduction of inconsistencies. The latter is very important
since bias problems have a big impact on the quantization error. Consequently we pay particular
attention in eliminating bias error in Section~\ref{subsec:quantized_inference}.

\subsection{Quantized inference}
\label{subsec:quantized_inference}
\begin{figure}
\centering
\includegraphics[width=\columnwidth]{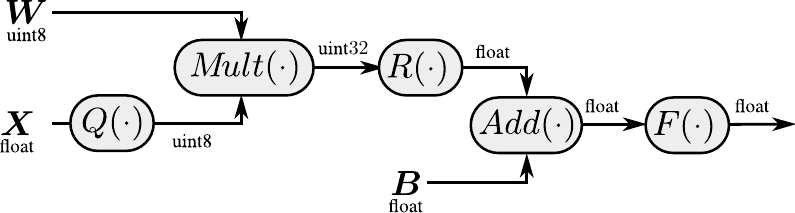}
\footnotesize
\caption{\footnotesize Execution of typical inference $y = W X + B$ : weights $\mathbf{W}$ are
already quantized, and inputs $\mathbf{X}$ are quantized $Q(\cdot)$ on-the-fly before
performing multiplication $Mult(\cdot)$; the product is then recovered $R(\cdot)$ to
apply the biases $\mathbf{B}$ and the activation function $F(\cdot)$.}
\label{fig:quantization_scheme_diagram}
\end{figure}

The approach we follow during neural network inference is to treat each layer independently,
receiving and producing floating point values: inputs get quantized on-the-fly, while network parameters
offline. Internally, layers operate on 8-bit integers for
the matrix multiplications (typically the most computationally intensive operations), and their
product is recovered to floating point, as depicted in Figure \ref{fig:quantization_scheme_diagram}.
This simplifies the implementation of complex activation functions, and allows mixing integer layers
with float layers, if desired.

\textbf{Multiplication of quantized values.}
In order to perform most of the multiplication,
$V_\text{c} = V_\text{a} * V_\text{b}$, in the 8-bit domain (though using 32-bit accumulators),
we must first apply the offset, $V_\text{min}$, such that
$V_\text{x} = \frac{V_\text{x}''}{Q}$, where  $V_\text{x}''=V_\text{x}' + Q V_\text{min}$.
We then apply the recovery factor on the result, which for the multiplication of
two independently quantized values $V_\text{a}'$ and $V_\text{b}'$, is the inverse
product of their quantization factors $Q_\text{a}$ and $Q_\text{b}$ :
\begin{equation}
\label{eq:mult}
\footnotesize
V_\text{c} = \frac{V_\text{a}'' * V_\text{b}''}{Q_\text{a} * Q_\text{b}}\quad\quad\quad\quad\quad\text{(} Mult(\cdot) \text{ and } R(\cdot) \text{ in Figure 1)}
\end{equation}


\textbf{Integer multiplication: effects on quantization and recovery.}
In eq.~\eqref{eq:mult}, each factor $V_\text{x}''$ is of integer type. This
means in the $V_\text{x}''$ formulation: 1) $V_\text{x}'$ is already an integer;
2) $Q V_\text{min}$ is a float that will be rounded to an integer, and thus
introduce an error: $E = \text{float}(Q V_\text{min}) - \text{integer}(Q V_\text{min})$.

This requires that quantization be performed in a way that is consistent
with this formulation in order to avoid introducing bias error. Thus we
introduce a rounding operation $\text{round}(\cdot)$:
\begin{equation}
\label{eq:quant}
\footnotesize
V_\text{x}' = \text{round}(Q V_\text{x}) - \text{round}(Q V_\text{min})\quad\quad\text{(} Q(\cdot) \text{ in Figure 1)}
\end{equation}

Thus precision errors in the quantization and multiplication are
consistent and cancel each other. This also means that recovery needs to
be consistent with eq.~\eqref{eq:quant}:
\begin{equation}
\footnotesize
V_\text{x} = \frac{V_\text{x}' + \text{round}(Q V_\text{min})}{Q}\quad\quad\quad\quad\quad\quad\text{(} R(\cdot) \text{ in Figure 1)}
\end{equation}

\textbf{Efficient implementation.} The proposed quantization scheme
benefits from the reduced memory bandwidth of accessing 8-bit values, and enables
squeezing more values into any fast cache available, thus reducing power consumption and
access time. Furthermore, it allows better use of optimized SIMD instructions
by fitting in more values per operation, which offers a performance advantage over their
floating point counterparts. The overhead of the quantization and recovery
operations is typically negligible, and also parallelizable via SIMD. We do not cover any
specific implementation since whereas the previous benefits are generally applicable, the
details hinge on the targeted hardware, and that is beyond the scope of this paper. However,
in our previous work ~\cite{mcgraw16} we recorded a significant speed up over unquantized
floating point inference.

Logically, our scheme can be applied at a given level of granularity, subdividing groups of
values into sub-groups for better precision. This means parameter matrices
at different NN layers can be quantized independently, or
even further broken down into individually quantized sub-matrices.
We set the granularity at the level of the weight
matrices (e.g. the parameters associated with individual gates in an LSTM).
This results in a relatively small loss in final inference accuracy
(see Table~\ref{tbl:results}).

\subsection{Quantization aware training}
\label{subsec:quantization-aware_training}
In order to minimize the loss from the quantization scheme described in
Section~\ref{subsec:quantized_inference}, we make it part of the
training process, under the principle that quantization noise must be considered
when computing the model's overall error, and thus gradients~\cite{KimS16}. We
maintain the full-precision (floating point) version of the parameters, but perform the
forward pass in quantized form as described in~\ref{subsec:quantized_inference},
thus mimicking what occurs during inference at run-time. The backward pass remains
in full-precision form, so that the gradient is also computed in full-precision
but is based on the error from the quantized forward pass. Thus the gradient
is used to update the full-precision parameters, which then in turn get
re-quantized to start a new forward pass. Unlike other
approaches~\cite{CourbariauxBD15}, we do not directly add the quantization component
during the backward pass since it is expected that the weights contribute in the same
proportions regardless of whether they are quantized or not. Moreover, we do not
want to introduce the accuracy error of the quantized operation when computing
the gradients. See Algorithm ~\ref{alg:quantization-aware_SGD}.

\begin{algorithm}
\footnotesize
\caption{\footnotesize Quantization aware SGD training. Where $L$ is the number of layers.
$C$ is the cost function, $\text{infer-and-recover}(\cdot)$ is a function that performs the
inference computation in integer form but returns the results in recovered
floating point. $\text{error}(\cdot),\text{wgradient}(\cdot),\text{bgradient}(\cdot),\text{adjust}(\cdot)$
are functions that perform the typical backpropagation operations in floating point.}
\label{alg:quantization-aware_SGD}
\begin{algorithmic}[1]
\Require{a mini-batch of (inputs, outputs), parameters $w_{t-1}$, and $b_{t-1}$
(weights and biases) in floating point precision, from previous training step
$t-1$.}
\Procedure{TrainingStep}{}
\State $w_{t-1}^q$ $\leftarrow$ quantize($w_{t-1}$)
\For{k=1 to L}
\State $a_k$  $\leftarrow$ infer-and-recover($a_{k-1}$, $w_{t-1}^q$, $b_{t-1}$)
\EndFor
\State Compute output error $\delta_L$
\For{k=L-1 to 2}
\State $\delta_k$ $\leftarrow$ error($w_{k+1,t-1}$, $\delta_{k+1}$, $a_{k+1}$)
\State $\frac{\partial C}{\partial w_{k, t-1}}$  $\leftarrow$ wgradient($a_{k-1}$, $delta_k$)
\State $\frac{\partial C}{\partial b_{k, t-1}}$  $\leftarrow$ bgradient($delta_k$)
\State $w_{k,t}$  $\leftarrow$ adjust($w_{t-1}$, $\frac{\partial C}{\partial w_{k, t-1}}$)
\State $b_{k,t}$  $\leftarrow$ adjust($b_{t-1}$, $\frac{\partial C}{\partial b_{k, t-1}}$)
\EndFor
\EndProcedure
\end{algorithmic}
\end{algorithm}

\begin{table*}
    \footnotesize
    \centering
    \begin{tabular}{|c|c|c|c|c||c|c|c|c|}
      \hline
      \textbf{System (Params.)} & \multicolumn{4}{|c||}{\textbf{WER (\%) on Clean Eval Set}} & \multicolumn{4}{|c|}{\textbf{WER (\%) on Noisy Eval Set}} \\
      \hline
        & \textbf{match} & \textbf{mismatch} & \textbf{quant} & \textbf{quant-all} & \textbf{match} & \textbf{mismatch} & \textbf{quant} & \textbf{quant-all} \\
      \hline
      \hline
      $4\times300$ ($\sim$2.9M) & 13.6 & 14.3 (5.1\%) & 13.5 (-0.7\%) & 13.6 (0.0\%) & 26.3 & 28.2 (7.2\%) & 26.5 (0.8\%) & 26.5 (0.8\%) \\
      \hline
      $5\times300$ ($\sim$3.7M) & 12.5 & 13.1 (4.8\%) & 12.6 (0.8\%) & 12.7 (1.6\%) & 24.6 & 26.6 (8.1\%) & 24.8 (0.8\%) & 25.0 (1.6\%) \\
      \hline
      $4\times400$ ($\sim$5.0M) & 12.1 & 12.5 (3.3\%)& 12.3 (1.7\%) & 12.3 (1.7\%) & 23.2 & 25.0 (7.8\%) & 23.7 (2.2\%) & 23.8 (2.6\%) \\
      \hline
      $5\times400$ ($\sim$6.3M) & 11.4 & 11.7 (2.6\%) & 11.5 (0.9\%) & 11.7 (2.6\%) & 22.3 & 23.5 (5.4\%) & 22.6 (1.3\%) & 22.7 (1.8\%) \\
      \hline
      $4\times500$ ($\sim$7.7M) & 11.7 & 12.0 (2.6\%) & 11.7 (0.0\%) & 11.7 (0.0\%) & 22.6 & 23.6 (4.4\%) & 22.6 (0.0\%)& 22.7 (0.4\%) \\
      \hline
      $5\times500$ ($\sim$9.7M) & 10.9 & 11.1 (1.8\%) & 11.2 (2.8\%) & 11.1 (1.8\%) & 20.9 & 21.7 (3.8\%) & 21.4 (2.4\%) & 21.5 (2.9\%) \\
      \hline
      \hline
      $P=100$ ($\sim$2.7M) & 11.6 & 12.1 (4.3\%) & 11.8 (1.7\%) & 11.9 (2.6\%) & 22.6 & 23.8 (5.3\%) & 23.1 (2.2\%) & 23.3 (3.1\%) \\
      \hline
      $P=200$ ($\sim$4.8M) & 10.6 & 10.8 (1.9\%) & 10.6 (0.0\%) & 10.7 (0.9\%) & 20.5 & 21.4 (4.4\%) & 20.6 (0.5\%) & 20.7 (1.0\%) \\
      \hline
      $P=300$ ($\sim$6.8M) & 10.3 & 10.5 (1.9\%) & 10.5 (1.9\%) & 10.6 (2.9\%) & 19.8 & 20.3 (2.5\%) & 20 (1.0\%) & 20.4 (3.0\%) \\
      \hline
      $P=400$ ($\sim$8.9M) & 10.3 & 10.5 (1.9\%) & 10.3 (0.0\%) & 10.5 (1.9\%) & 19.6 & 20.2 (3.1\%) & 19.8 (1.0\%) & 19.9 (1.5\%) \\
      \hline
      \hline
      Avg. Relative Loss & - & 3.0\% & 0.9\% & 1.6\% & - & 5.2\% & 1.2\% & 1.9\% \\
      \hline
    \end{tabular}
    \caption{\footnotesize Word error rates on `clean' and `noisy' evaluation sets for various
    model architectures. Numbers in parentheses represent the loss relative to
    the \textbf{`matched'} condition where models are trained and evaluated using
    floating point arithmetic.}
    \label{tbl:results}
\end{table*}


  \section{Experimental Setup}
  \label{sec:experiments-setup}
The focus of our experiments is to determine the impact of quantization in the
context of building small, efficient acoustic models on an open-ended
large-vocabulary speech recognition task. Following our previous
work~\cite{mcgraw16, prabhavalkar2016}, all models are trained to optimize the
connectionist temporal classification (CTC) loss
function~\cite{GravesFernandesGomezEtAl06}, followed by sequence
discriminative training to optimize the state-level minimum Bayes risk (sMBR)
criterion~\cite{Kingsbury09}.

We evaluate architectures which vary along two main dimensions: the total number
of parameters in the model, and whether the architecture uses projection
layers~\cite{SakSeniorBeaufays14} or not.
We train RNN-based acoustic models with 4 or 5 layers of LSTM cells; the number
of LSTM cells, $N$, is kept the same in all of the layers. We use $N=300,
400, 500$ in our experiments for a total of 6 configurations. In addition, we
train models with 5 layers of 500 LSTM cells, but insert a projection
layer of $P$ units after each of the 5 LSTM layers to reduce
the rank of the recurrent and inter-layer weight matrices~\cite{SakSeniorBeaufays14}.
In this work, unlike our previous work~\cite{prabhavalkar2016}, we keep the
size of the projection layers the same across all layers. We consider
$P=100, 200, 300, 400$, thus adding 4 more configurations.

We utilize the same frontend as our previous work~\cite{prabhavalkar2016}:
standard 40-dimensional log mel-filterbank energies over the 8kHz range,
computed every 10ms on 25ms windows of input speech.
Following~\cite{SakSeniorRaoEtAl15b}, we stack features together from 8
consecutive frames (7 frames of right context) and only present every third
stacked frame as input to the network. In addition to stabilizing CTC training,
this reduces computation since the network is only evaluated once every 30ms.
In order to minimize the delay between the acoustics and the output labels produced
by the network, we constrain the set of CTC alignments to be within 100ms of the
locations determined by a forced-alignment~\cite{Senior15}.
Our decoding setup is identical to that presented in our previous
work~\cite{mcgraw16, prabhavalkar2016}.
Following~\cite{lei2013}, we generate a much smaller first-pass language model (LM)
(69.5K n-grams; mostly unigrams) which is composed with the lexicon transducer to
generate the decoder graph; models are re-scored on-the-fly with a larger
5-gram LM.
Our systems are trained on anonymized hand-transcribed utterances extracted
from Google voice-search ($\sim$3M utterances) and dictation ($\sim$1M
utterances) traffic. To improve robustness, we create `multi-style' training
data by synthetically distorting the utterances, simulating the effect of
background noise and reverberation. 20 distorted utterances are created for
each input utterance; noise samples used in this process are extracted from
environmental recordings of everyday events and Youtube videos. Results are
reported on a set of 13.3K hand-transcribed anonymized utterances (135K words)
extracted from Google traffic from an open-ended dictation domain. We also
report results on a `noisy' version of the evaluation set, created
synthetically using a noise distribution with similar characteristics as the
one used to train the model.

\section{Experiments}
\label{sec:experiments}
In pilot experiments, we found that quantization aware CTC training did not produce
models with a better word error rate (WER) performance than `standard' float trained models.
Therefore, in all of our experiments we use float CTC training, and then
apply quantization aware sMBR training.

\subsection{CTC Training of LSTM AMs with Projection Layers}
\begin{figure}
\centering
\includegraphics[width=0.92\columnwidth]{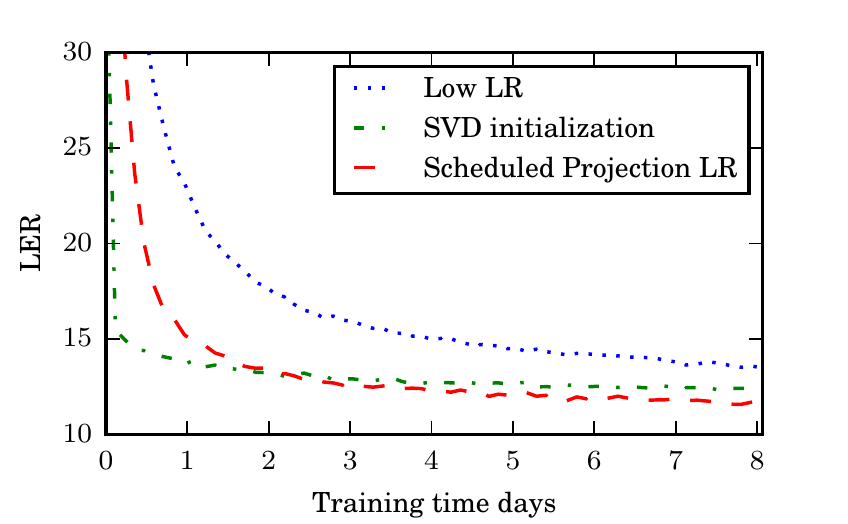}
\caption{\footnotesize Label error rates on held-out set as a function of training time
during CTC training of model with 200 projection layer nodes
$(P = 200)$ using different learning rate schedules.}
\label{fig:proj_layer_ler}
\end{figure}
As reported in our previous work~\cite{prabhavalkar2016}, we find that
training LSTMs to optimize the CTC criterion is somewhat unstable for models
with projection layers. One solution to this problem, which we proposed
in~\cite{prabhavalkar2016}, is to first train an `uncompressed' model without
any projection layers. This model is used to initialize the projection layer
matrices through a truncated singular value decomposition (SVD) of the
recurrent weight matrices. While this stabilizes the training
process, and has the benefit of providing a principled procedure for setting
the number of nodes in each of the projection layers, it has the drawback
that it requires a two-stage training process that increases
overall training time. Therefore, in the present work, we propose an alternative
strategy that stabilizes CTC training without requiring an expensive two-stage training
process, yet results in better convergence. As a representative example,
in Figure~\ref{fig:proj_layer_ler}, we plot CI-phoneme label error rates (LERs)
for the model with 200 projection nodes in
each layer $(P = 200)$, on a held-out development set. In all cases, we use an
exponentially decaying global learning rate (LR): $\eta_g(t) = c_g 10^{-
\frac{t}{T_g}}$, where $t$ is the total training time ($c_g = 1.5 \times
10^{-4}$ and $T_g = 20 \text{ days}$ in our experiments). The SVD-based
initialization~\cite{prabhavalkar2016} appears as `SVD initialization' in the
figure.

The most straightforward technique to stabilize training is to set the initial
global learning rate as high as possible while avoiding divergence
($c_g = 1.5 \times 10^{-7}$, in our experiments; `Low LR' in
Figure~\ref{fig:proj_layer_ler}). Although this stabilizes training, this leads
to extremely slow convergence since the learning
rate is many orders of magnitude smaller, and significantly worse LER than the
SVD-based initialization.

As an alternative, we propose using a lower learning rate for parameters in
the projection layer, by defining a separate projection learning rate
multiplier $\eta_p(t)$ which multiplies the global learning rate (i.e., the effective learning rate
for projection layer parameters is $\eta_g(t)\eta_p(t)$). We find that we can
stabilize training by using a lower effective learning rate for the projection
layer parameters relative to the rest of the system by using an exponentially
increasing projection learning rate multiplier that gradually scales the
effective learning rate multipler towards the global learning rate (`Scheduled
Projection LR' in Figure~\ref{fig:proj_layer_ler}): $\eta_{p}(t) = c_{p}^{\left(1 -
\min\left\{\frac{t}{T_{p}}, 1\right\}\right)}$ ($c_p = 10^{-3}$ and $T_p = 0.6 \text{
days}$ in our experiments). Note that, $\eta_p(t) \rightarrow 1 \text{ as } t
\rightarrow T_p$, and thus the same effective learning rate is used for all
parameters for $t > T_p$. As can be seen in Figure~\ref{fig:proj_layer_ler},
although the SVD-based initialization outperforms using a single low global
learning rate, the scheduled projection learning rate schedule results in the
fastest convergence, while avoiding the need for the two-stage training
required by the SVD-based initialization. Therefore, we employ the projection
learning rate schedule for CTC training of models with projection layers.
\subsection{Quantization aware sMBR Training of AMs}
\label{sec:am-smbr-qtrain}

Once models have been trained under the CTC criterion, we
sequence-train them to optimize the sMBR criterion.
In order to mitigate the instability encountered during sMBR
training of models with projection layers, we find that it is sufficient to
use a constant learning rate multiplier for projection layer nodes:
$\eta_p(t) = c^{\text{sMBR}}_{p}$ (we set,
$c^{\text{sMBR}}_{p} = 0.5$ and the global LR parameter,
$c_g = 1.5\times10^{-5}$, in our experiments).

  \section{Results}
  \label{sec:results}
We report results separately for `clean' and `noisy' evaluation conditions
in Table~\ref{tbl:results}. Our baseline models are trained using `standard'
floating-point arithmetic during forward and backward passes through the
network. We evaluate the float-trained baseline models with quantization applied
only after training (\textbf{`mismatch'} in Table~\ref{tbl:results}), and with
floating-point precision, i.e., \emph{without quantization}
(\textbf{`match'} in Table~\ref{tbl:results}). The latter allows measuring the
loss introduced by quantization, and represents a ceiling WER
performance on our task.
We examine the impact of the proposed quantization aware training
using two schemes during training time: quantizing all layers except for
the final softmax layer (\textbf{`quant'} in Table~\ref{tbl:results}), or
quantizing all layers in the network (\textbf{`quant-all'} in
Table~\ref{tbl:results}). In both cases evaluation occurs in quantized form, as
in \textbf{`mismatch'}.

Comparing performance obtained in the \textbf{match} and \textbf{mismatch}
configurations in Table~\ref{tbl:results}, we observe that the
relative loss in accuracy due to quantization is greater on the noisy evaluation set
(up to 8.1\%) than in the clean evaluation set (up to 5.1\%). In general,
the loss due to quantization is inversely proportional to the number of
parameters in the model. It is also interesting to compare WER performance obtained
using the two kinds of model architectures examined in this work. We note that
models which employ projection layers appear to outperform similarly sized
models without projection layers (cf., $P=200$ compared to models with 4 or 5
layers of 400 cells ($4 \times 400$ or $5 \times 400$)). Models
with projection layers appear to suffer less degradation in performance after
quantization, thus making them desirable for resource-constrained
speech recognition tasks~\cite{mcgraw16,prabhavalkar2016}.

The proposed quantization aware training scheme significantly improves
performance relative to the \textbf{`mismatch'} condition; in some cases the
proposed technique recovers all of the loss due to quantization
(cf., clean set for $4 \times 300$ and $4 \times 500$). Quantization aware
training results in gains on both evaluation sets, with larger
gains obtained on the noisy set. We note that quantizing all layers, except
for the final softmax layer (\textbf{`quant'}) appears to be slightly superior to
quantizing all layers during training (\textbf{`quant-all'}).
When comparing the relative loss in performance due to quantization, averaged across
all of the model architectures, quantization aware training recovers 2.1\% of the
loss on the clean evaluation set, and 4\% of the loss on the noisy evaluation set.

  \section{Conclusions}
  \label{sec:conclusions}
We propose a simple and computationally efficient quantization scheme for training and execution of deep learning models. It allows us to reduce the resolution of weights in neural networks from 32-bit floating point values to 8-bit integers, thus enabling fast inference by means of optimized hardware operations and reduced memory bandwidth. We also utilize the quantization scheme during training to recover the accuracy loss caused by quantization noise during inference. In experimental evaluations we find that our quantization scheme, when applied only \emph{after} training, results in moderate loss in recognition quality. We also present how such loss is almost completely eliminated by quantization aware training.

  \newpage
  \eightpt
  \bibliographystyle{IEEEtran}

  \bibliography{refs}

\end{document}